\pdfoutput=1

\documentclass[11pt]{article}

\usepackage[final]{EMNLP2023}

\usepackage{times}
\usepackage{latexsym}
\usepackage{amsmath}
\usepackage{enumitem}
\usepackage{graphicx}
\usepackage{tikz}
\usetikzlibrary{patterns}
\usetikzlibrary{matrix}
\usepackage{pgfplots}
\usepgfplotslibrary{groupplots}
\usepackage{inconsolata}
\usepackage{booktabs}
\usepackage{multirow}
\usepackage{hyperref}
\usepackage{comment}
\pgfplotsset{compat=1.7}
\definecolor{colorone}{RGB}{120,200,192}
\definecolor{colortwo}{RGB}{199,234,229}
\definecolor{colorthree}{RGB}{1,102,94}

\definecolor{colorwayone}{RGB}{166,97,26}
\definecolor{colorwaytwo}{RGB}{223,194,125}
\definecolor{colorwaythree}{RGB}{245,245,245}
\definecolor{colorwayfour}{RGB}{128,205,193}
\definecolor{colorwayfive}{RGB}{1,133,113}

\usepackage[T1]{fontenc}

\usepackage[utf8]{inputenc}

\usepackage{microtype}
\usepackage{amsmath}

\usepackage{inconsolata}
\usepackage{xspace}

\definecolor{myblue}{rgb}{0.9, 0.1, 0.94}
\definecolor{mygreen}{rgb}{0.64, 0.56, 0.88}
\definecolor{myyellow}{rgb}{0.98, 0.94, 0.75}
\definecolor{mygreen}{rgb}{0, 1, 0}

\newcommand{\std}[1]{\color{gray}{\xspace#1\xspace}}
\newcommand{\emptystd}[1]{\color{white}{\xspace#1\xspace}}

\title{Large Language Models Enable Few-Shot Clustering}

\author{
\bf{Vijay Viswanathan}\textsuperscript{1}, \bf{Kiril Gashteovski}\textsuperscript{2}, \\ \bf{Carolin Lawrence}\textsuperscript{2}, \bf{~Tongshuang Wu}\textsuperscript{1}, \bf{Graham Neubig}\textsuperscript{1, 3} \\ 
\textsuperscript{1} Carnegie Mellon University, \textsuperscript{2} NEC Laboratories Europe, \textsuperscript{3} Inspired Cognition }

\begin{document}
\maketitle
\begin{abstract}
Unlike traditional unsupervised clustering, semi-supervised clustering allows \emph{users} to provide meaningful structure to the data, which helps the clustering algorithm to match the user's intent. Existing approaches to semi-supervised clustering require a significant amount of feedback from an expert to improve the clusters. In this paper, we ask whether a large language model can \emph{amplify} an expert's guidance to enable query-efficient, \emph{few-shot} semi-supervised text clustering.
We show that LLMs are surprisingly effective at improving clustering. We explore three stages where LLMs can be incorporated into clustering: before clustering (improving input features), during clustering (by providing constraints to the clusterer), and after clustering (using LLMs post-correction). We find incorporating LLMs in the first two stages can routinely provide significant improvements in cluster quality, and that LLMs enable a user to make trade-offs between cost and accuracy to produce desired clusters. We release our code and LLM prompts for the public to use.\footnote{\url{https://github.com/viswavi/few-shot-clustering}}
\end{abstract}

\section{Introduction}
Unsupervised clustering aims to do an impossible task: organize data in a way that satisfies a domain expert's needs without any specification of what those needs are. Clustering, by its nature, is fundamentally an \textit{underspecified} problem. According to \citet{caruana_pau}, this underspecification makes clustering ``probably approximately useless.''

Semi-supervised clustering, on the other hand, aims to solve this problem by enabling the domain expert to guide the clustering algorithm \citep{interactive_clustering_survey}. Prior works have introduced different types of interaction between an expert and a clustering algorithm, such as initializing clusters with hand-picked seed points \citep{Basu2002SemisupervisedCB}, specifying pairwise constraints \citep{Basu2004ActiveSF, Zhang2019AFF}, providing feature feedback \citep{Dasgupta2010WhichCD}, splitting or merging clusters \citep{Awasthi2013LocalAF}, or locking one cluster and refining the rest \citep{Coden2017AMT}. These interfaces have all been shown to give experts control of the final clusters. However, they require significant effort from the expert. For example, in a simulation that uses split/merge, pairwise constraint, and lock/refine interactions \cite{Coden2017AMT}, it took between 20 and 100 human-machine interactions to get \emph{any} clustering algorithm to produce clusters that fit the human's needs. Therefore, for large, real-world datasets with a large number of possible clusters, the feedback cost required by interactive clustering algorithms can be immense.

\begin{figure}[t]
\begin{center}

    \includegraphics[width=0.45\textwidth]{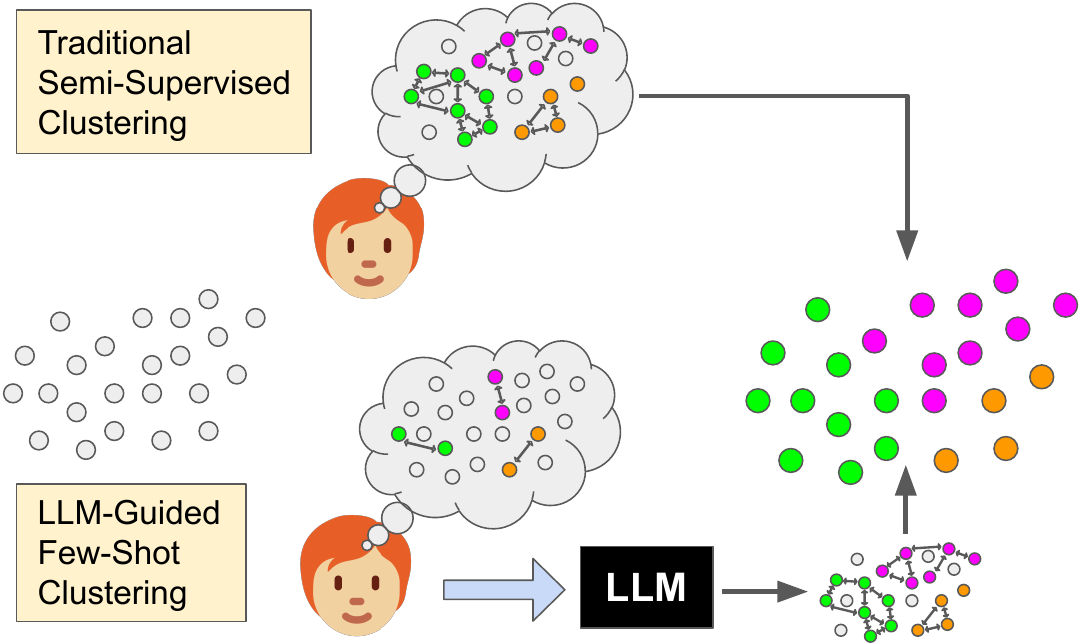}
    
\end{center}
\caption{In traditional semi-supervised clustering, a user provides a large amount of feedback to the clusterer. In our approach, the user prompts an LLM with a small amount of feedback. The LLM then generates a large amount of pseudo-feedback for the clusterer.
}
\label{fig:intro_diagram}
\vspace{-10pt}
\end{figure}

Building on a body of recent work that uses Large Language Models (LLMs) as noisy simulations of human decision-making \citep{Fu2023GPTScoreEA, NBERw31122, park2023generative}, we propose a different approach for semi-supervised text clustering. In particular, we answer the following research question: \emph{Can an expert provide a few demonstrations of their desired interaction (e.g., pairwise constraints) to a large language model, then let the LLM direct the clustering algorithm?} 

We explore three places in the text clustering process where an LLM could be leveraged: before clustering, during clustering, and after clustering.
We leverage an LLM \emph{before clustering} by augmenting the textual representation. For each example, we generate keyphrases with an LLM, encode these keyphrases, and add them to the base representation.  We incorporate an LLM \emph{during clustering} by adding cluster constraints. Adopting a classical algorithm for semi-supervised clustering, pairwise constraint clustering, we use an LLM as a pairwise constraint pseudo-oracle. We then explore using an LLM \emph{after clustering} by correcting low-confidence cluster assignments using the pairwise constraint pseudo-oracle. In every case, the interaction between a user and the clustering algorithm is enabled by a prompt written by the user and provided to a large language model.

We test these three methods on five datasets across three tasks: canonicalizing entities, clustering queries by intent, and grouping tweets by topic.
We find that, compared to traditional K-Means clustering on document embeddings, using an LLM to enrich each document's representation empirically improves cluster quality on every metric for all datasets we consider. Using an LLM as a pairwise constraint pseudo-oracle can also be highly effective when the LLM is capable of providing pairwise similarity judgements but requires a larger number of LLM queries to be effective. However, LLM post-correction provides limited upside.
Importantly, LLMs can also approach the performance of \emph{traditional semi-supervised clustering with a human oracle} at a fraction of the cost. 

Our work stands out from recent deep-learning-based text clustering methods~\cite{SCCL, Zhang2023ClusterLLMLL} in its remarkable simplicity. Using an LLM to expand documents' representation or correct clustering outputs can be added as a plug-in to \textit{any text clustering algorithm} using \textit{any set of text features}, while our pseudo-oracle pairwise constraint clustering approach requires using K-Means as the underlying clustering algorithm.
In our investigation of what aspect of the LLM prompt is most responsible for the clustering behavior, we find that just using an instruction alone (with no demonstrations) adds significant value. This can motivate future research directions for integrating natural language instructions with a clustering algorithm.



\section{Methods to Incorporate LLMs}
\label{sec:methods}

In this section, we describe the methods that we use to incorporate LLMs into clustering.

\subsection{Clustering via LLM Keyphrase Expansion}
Before any cluster is produced, experts typically know what aspects of each document they wish to capture during clustering.
Instead of forcing clustering algorithms to mine such key factors from scratch, it could be valuable to globally highlight these aspects (and thereby specify the task emphases) beforehand. 
To do so, we use an LLM to make every document's textual representation \emph{task-dependent}, by enriching and expanding it with evidence relevant to the clustering need. 
Specifically, each document is passed through an LLM which generates keyphrases, these keyphrases are encoded by an embedding model, and the keyphrase embedding is then concatenated to the original document embedding.

\begin{figure}[h]
\begin{center}

    \includegraphics[width=0.47\textwidth]{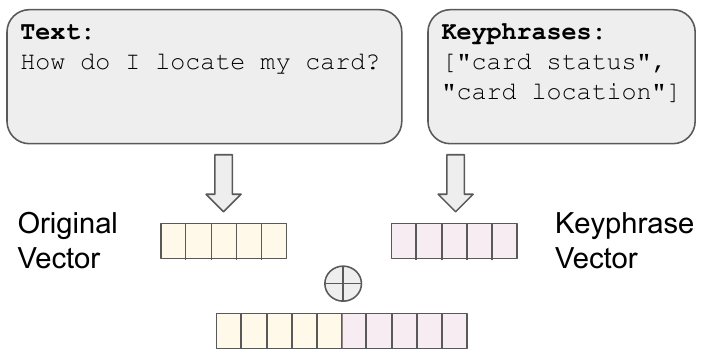}
    
\end{center}
\vspace{-6pt}
\caption{We expand document representations by concatenating them with keyphrase embeddings. The keyphrases are generated by a large language model.}
\label{fig:keyphrase_diagram}
\vspace{-10pt}
\end{figure}

We generate keyphrases using GPT-3 (specifically, \texttt{gpt-3.5-turbo-0301}). We provide a short prompt to the LLM, starting with an instruction (e.g. \emph{``I am trying to cluster online banking queries based on whether they express the same intent. For each query, generate a comprehensive set of keyphrases that could describe its intent, as a JSON-formatted list.''}). The instruction is followed by four demonstrations of keyphrases (example shown in \autoref{fig:keyphrase_diagram}). Examples of full prompts are shown in \autoref{sec:keyphrase_expansion_prompt}.

We then encode the generated keyphrases into a single vector, and concatenate this vector with the original document's text representation. To disentangle the knowledge from an LLM with the benefits of a better encoder, we encode the keyphrases using the same encoder as the original text.\footnote{An exception to this is entity clustering. There, the BERT encoder has been specialized for clustering Wikipedia sentences, so we use DistilBERT to support keyphrase clustering.}

\subsection{Pseudo-Oracle Pairwise Constraint Clustering}
\label{sec:pckmeans_method}
We explore the situation where a user conceptually describes which kinds of points to group together and wants to ensure  the final clusters follow this grouping.

Arguably, the most popular approach to semi-supervised clustering is \emph{pairwise constraint clustering}, where an oracle (e.g. a domain expert) selects pairs of points which \textit{must} be linked or \textit{cannot} be linked \citep{Wagstaff2000ClusteringWI}, such that more abstract clustering needs of experts can be implicitly induced from the concrete feedback.

We use this paradigm to investigate the potential of LLMs to amplify expert guidance during clustering, using an LLM as a \textit{pseudo-oracle}.

To select pairs to classify, we take different strategies for entity canonicalization and for other text clustering tasks.  For text clustering, we adapt the Explore-Consolidate algorithm \citep{Basu2004ActiveSF} to first collect a diverse set of pairs from embedding space (to identify pairs of points that must be linked), then collect points that are nearby to already-chosen points (to find pairs of points that cannot be linked). For entity canonicalization, where there are so many clusters that very few pairs of points must be linked, we simply identify the closest distinct pairs of points in embedding space.

We prompt an LLM with a brief domain-specific instruction (provided in entirety in \autoref{sec:pairwise_constraint_prompt}), followed by up to 4 demonstrations of pairwise constraints, obtained from test set labels. We use these pairwise constraints to generate clusters with the PCKMeans algorithm of \citet{Basu2004ActiveSF}. This algorithm applies penalties for cluster assignments that violate any constraints, weighted by a hyperparameter $w$. Following prior work \cite{vashishth2018cesi}, we tune this parameter on each dataset's validation split.

\subsection{Using an LLM to Correct a Clustering}
\label{sec:kmeans_correction}
\begin{figure}[t]
\begin{center}

    \includegraphics[width=0.35\textwidth]{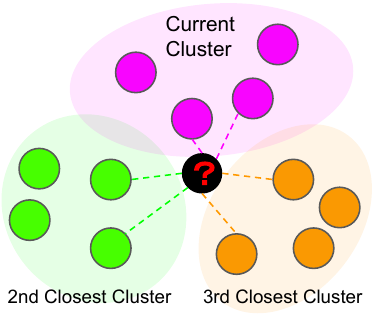}
    
\end{center}
\vspace{-6pt}
\caption{After performing clustering, we identify low-confidence points. For these points, we ask an LLM whether the current cluster assignment is correct. If the LLM responds negatively, we ask the LLM whether this point should instead be linked to any of the top-5 nearest clusters, and correct the clustering accordingly.}
\label{fig:kmeans_correction_diagram}
\vspace{-10pt}
\end{figure}

We finally consider the setting where one has an existing set of clusters, but wants to improve their quality with minimal local changes. 
We use the same pairwise constraint pseudo-oracle as in \autoref{sec:pckmeans_method} to achieve this, and we illustrate this procedure in \autoref{fig:kmeans_correction_diagram}.

We identify the \emph{low-confidence points} by finding the $k$ points with the least margin between the nearest and second-nearest clusters (setting $k=500$ for our experiments).
We textually represent each cluster by the entities nearest to the centroid of that cluster in embedding space. For each low-confidence point, we first ask the LLM whether or not this point is correctly linked to any of the representative points in its currently assigned cluster. If the LLM predicts that this point should not be linked to the current cluster, we consider the 4 next-closest clusters in embedding space as candidates for reranking, sorted by proximity.
To rerank the current point, we ask the LLM whether this point should be linked to the representative points in each candidate cluster. If the LLM responds positively, then we reassign the point to this new cluster. If the LLM responds negatively for all alternative choices, we maintain the existing cluster assignment.

\section{Tasks}

\subsection{Entity Canonicalization}

\paragraph{Task.}
In \textit{entity canonicalization}, we must group a collection of noun phrases $M = \{m_i\}_1^N$ into subgroups $\{C_j\}_1^K$ such that $m_1 \in C_j$ and $m_2 \in C_j$ if and only if $m_1$ and $m_2$ refer to the same entity. For example, the noun phrases \emph{President Biden ($m_1$), Joe Biden ($m_2$)} and \emph{the 46th U.S.~President ($m_3$)} should be clustered in one group (e.g., $C_1$). The set of noun phrases $M$ are usually the nodes of an ``open knowledge graph'' produced by an OIE system.\footnote{Open Information Extraction (OIE) is the task of extracting surface-form \emph{(subject; relation; object)}-triples from natural language text in a schema-free manner \cite{Banko2007OpenIE}.} Unlike the related task of entity linking \citep{Bunescu2006UsingEK, Milne2008LearningTL}, we do not assume that any curated knowledge graph, gazetteer, or encyclopedia contains all the entities of interests.

Entity canonicalization is valuable for motivating the challenges of semi-supervised clustering. Here, there are hundreds or thousands of clusters and relatively few points per cluster, making this a difficult clustering task that requires lots of human feedback to be effective.

\paragraph{Datasets.}
We experiment with two datasets:

\begin{itemize}[nosep, leftmargin=1.2em,labelwidth=*,align=left]
\item \textit{OPIEC59k} \citep{CMVC} contains 22K noun phrases (with 2138 unique entity surface forms) belonging to 490 ground truth clusters. The noun phrases are extracted by MinIE \citep{Gashteovski2017MinIEMF, gashteovski2019opiec}, and the ground truth entity clusters are anchor texts from Wikipedia that link to the same Wikipedia article.

\item \textit{ReVerb45k} \citep{vashishth2018cesi} contains 15.5K mentions (with 12295 unique entity surface forms) belonging to 6700 ground truth clusters. The noun phrases are the output of the ReVerb \citep{Fader2011IdentifyingRF} system, and the ``ground-truth'' entity clusters come from automatically linking entities to the Freebase knowledge graph. We use the version of this dataset from \citet{CMVC}, who manually removed samples containing labeling errors.
\end{itemize}



\paragraph{Canonicalization Metrics.}
We follow the standard metrics used by \citet{CMVC}:
\begin{itemize}[nosep, leftmargin=1.2em,labelwidth=*,align=left]
    \item \textit{Macro Precision and Recall}

    \begin{itemize}[nosep, leftmargin=1.2em,labelwidth=*,align=left]
    \item Prec: For what fraction of predicted clusters is every element in the same gold cluster?
    \item Rec: For what fraction of gold clusters is every element in the same predicted cluster?
    \end{itemize}

\item \textit{Micro Precision and Recall}
    \begin{itemize}[nosep, leftmargin=1.2em,labelwidth=*,align=left]
    \item Prec: How many points are in the same gold cluster as the majority of their predicted cluster?
    \item Rec: How many points are in the same predicted cluster as the majority of their gold cluster?
    \end{itemize}

\item \textit{Pairwise Precision and Recall}
    \begin{itemize}[nosep, leftmargin=1.2em,labelwidth=*,align=left]
    \item Prec: How many pairs of points predicted to be linked are truly linked by a gold cluster?
    \item Rec: How many pairs of points linked by a gold cluster are also predicted to be linked?
    \end{itemize}

\end{itemize}
\noindent We finally compute the harmonic mean of each pair to obtain \textit{Macro F1}, \textit{Micro F1}, and \textit{Pairwise F1}.

\subsection{Text Clustering}

\paragraph{Task.}
We then consider the case of clustering short textual documents. This clustering task has been extensively studied in the literature \citep{Aggarwal2012ASO}.

\paragraph{Datasets.}
We use three datasets in this setting:
\begin{itemize}[nosep, leftmargin=1.2em,labelwidth=*,align=left]
\item \textit{Bank77} \citep{casanueva-etal-2020-efficient} contains 3,080 user queries for an online banking assistant from 77 intent categories.

\item \textit{CLINC} \citep{larson-etal-2019-evaluation} contains 4,500 user queries for a task-oriented dialog system from 150 intent categories, after removing ``out-of-scope'' queries (as in \citep{Zhang2023ClusterLLMLL}.

\item \textit{Tweet} \citep{Yin2016AMA} contains 2,472 tweets from 89 categories.
\end{itemize}
\paragraph{Metrics.}
Following prior work \citep{SCCL}, we compare our text clusters to the ground truth using normalized mutual information and accuracy (obtained by finding the best alignment between ground truth and predicted clusters using the Hungarian algorithm \citep{Hungarian}).

\section{Baselines}
\label{sec:baseline}
\subsection{K-Means on Embeddings}
We build our methods on top of a baseline of K-means clustering \citep{kmeans_clustering} over encoded data with k-means++ cluster initialization \citep{kpp}.
We choose the features and number of cluster centers that we use by task, largely following previous work.

\looseness=-1
\paragraph{Entity Canonicalization}

\begin{figure}[t]
\begin{center}

    \includegraphics[width=0.4\textwidth]{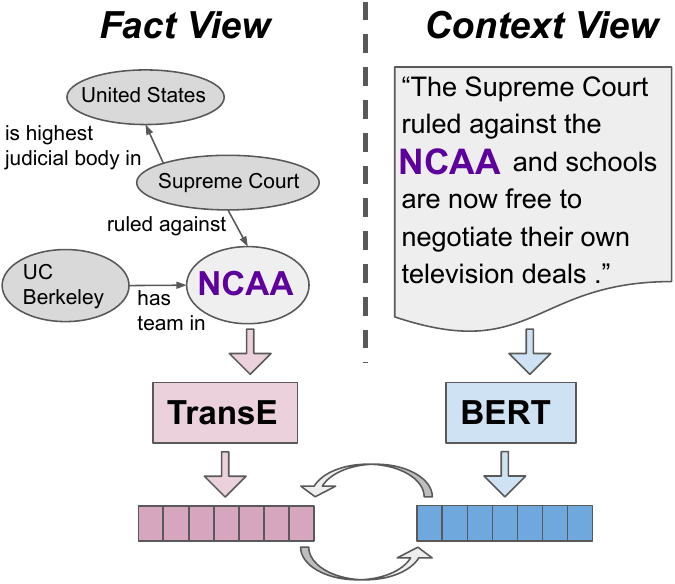}
    
\end{center}
\caption{Using the CMVC architecture, we encode a knowledge graph-based ``fact view'' and a text-based ``context-view'' to represent each entity.
}
\label{fig:cmvc_architecture}
\end{figure}

Following prior work \citep{vashishth2018cesi, CMVC}, we cluster individual entity mentions (e.g. ``ever since the ancient Greeks founded the city of \textit{Marseille} in 600 BC.'') by representing unique surface forms (e.g. ``Marseille'') globally, irrespective of their particular mention context. After clustering unique surface forms, we compose this cluster mapping onto the individual mentions (extracted from individual sentences) to obtain mention-level clusters.

We build off of the ``multi-view clustering'' approach of \citet{CMVC}, and represent each noun phrase using textual mentions from the Internet and the ``open'' knowledge graph extracted from an OIE system, as shown in \autoref{fig:cmvc_architecture}. They use a BERT encoder \citep{bert} to represent the textual context where an entity occurs (called the ``context view''), and a TransE knowledge graph encoder \citep{transe} to represent nodes in the open knowledge graph (called the ``fact view''). They improve these encoders by finetuning the BERT encoder using weak supervision of coreferent entities and improving the knowledge graph representations using data augmentation on the knowledge graph. These two views of each entity are then combined to produce a representation.

In their original paper, they propose an alternating multi-view K-Means procedure where cluster assignments that are computed in one view are used to initialize cluster centroids in the other view. After a certain number of iterations, if the per-view clusterings do not agree, they perform a ``conflict resolution'' procedure to find a final clustering with low inertia in both views. 
One of our secondary contributions is a simplification of this algorithm. We find that by simply using their finetuned encoders, concatenating the representations from each view, and performing K-Means clustering with K-Means++ initialization \citep{kpp} in a shared vector space, we can match their reported performance.

Finally, regarding the number of cluster centers, following the Log-Jump method of \citet{CMVC}, we choose 490 and 6687 clusters for OPIEC59k and ReVerb45k, respectively.

\paragraph{Intent Clustering}
For the Bank77 and CLINC datasets, we follow \citet{Zhang2023ClusterLLMLL} and encode each user query using the Instructor encoder. We use a simple prompt to guide the encoder: ``Represent utterances for intent classification''.
Again following previous work, we choose 150 and 77 clusters for CLINC and Bank77, respectively.

\paragraph{Tweet Clustering}
Following \citet{SCCL}, we encode each tweet using a version of DistilBERT \citep{Sanh2019DistilBERTAD} finetuned for sentence similarity classification\footnote{This model is \texttt{distilbert-base-nli-stsb-mean-tokens} on HuggingFace.} \citep{reimers-2019-sentence-bert}.
We use 89 clusters \citep{SCCL}.

\subsection{Clustering via Contrastive Learning}
In addition to the methods described in Section \ref{sec:methods}, we also include two other methods for text clustering, where previously reported: SCCL \citep{SCCL} and ClusterLLM \citep{Zhang2023ClusterLLMLL}. Both use constrastive learning of deep encoders to improve clusters, making these significantly more complicated and compute-intensive than our proposed methods. SCCL combines deep embedding clustering \citep{Xie2015UnsupervisedDE} with unsupervised contrastive learning to learn features from text. ClusterLLM uses LLMs to improve the learned features. After running hierarchical clustering, they also use triplet feedback from the LLM (``is point A more similar to point B or point C?'') to decide the cluster granularity from the cluster hierarchy and generate a flat set of clusters. To compare effectively with these approaches, we use the same encoders reported for SCCL and ClusterLLM in prior works: Instructor \citep{INSTRUCTOR} for Bank77 and CLINC and DistilBERT (finetuned for sentence similarity classification)  \citep{Sanh2019DistilBERTAD, reimers-2019-sentence-bert} for Tweet.

\begin{table*}[t] 
  \centering
  \renewcommand{\arraystretch}{1.1}
  \fontsize{9}{9.5}\selectfont
  \setlength{\tabcolsep}{5pt}
  \begin{tabular}{@{} lr|rrrr | rrrr @{} }
  \toprule
  \toprule
   &  \multirow{2}{*}{Dataset / Method} & \multicolumn{4}{c|}{OPIEC59k} & \multicolumn{4}{c}{ReVerb45k} \\
   \cmidrule(lr){3-6} \cmidrule(lr){7-10}
   &  & \multicolumn{1}{c}{\textbf{Macro F1}} & \multicolumn{1}{c}{\textbf{Micro F1}} & \multicolumn{1}{c}{\textbf{Pair F1}} & \multicolumn{1}{c|}{\textbf{\emph{Avg}}}   & \multicolumn{1}{c}{\textbf{Macro F1}} & \multicolumn{1}{c}{\textbf{Micro F1}} & \multicolumn{1}{c}{\textbf{Pair F1}} & \multicolumn{1}{c}{\textbf{\emph{Avg}}}  \\ 
   \midrule\midrule
  & Optimal Clust.  & 80.3\emptystd{$\pm$0.0} & 97.0\emptystd{$\pm$0.0} & 95.5\emptystd{$\pm$0.0} & 90.9 
     & 84.8\emptystd{$\pm$0.0} & 93.5\emptystd{$\pm$0.0} & 92.1\emptystd{$\pm$0.0} & 90.1\\
  \midrule
  & CMVC  & 52.8\emptystd{$\pm$0.0} & 90.7\emptystd{$\pm$0.0} & 84.7\emptystd{$\pm$0.0} & 76.1 
    & 66.1\emptystd{$\pm$0.0} & 87.9\emptystd{$\pm$0.0} & 89.4\emptystd{$\pm$0.0} & 81.1 \\
  \midrule
  & KMeans  & 53.5\std{$\pm$0.0} & 91.0\std{$\pm$0.0} & 85.6\std{$\pm$0.0} & 76.7 & 69.6\std{$\pm$0.0} & 89.1\std{$\pm$0.0} & 89.3\std{$\pm$0.0} & 82.7\\
  \midrule

  \parbox[t]{1.5mm}{\multirow{3}{*}{\rotatebox[origin=c]{90}{\begin{small}ours\end{small}}}} 
  & PCKMeans & 58.7\std{$\pm$0.0} & 91.5\std{$\pm$0.0} & 86.1\std{$\pm$0.0} & 78.7 
  & 72.0\std{$\pm$0.0} & 88.5\std{$\pm$0.0} &   87.0\std{$\pm$0.0}  & 82.5\\
  & LLM Correction  & 58.7\emptystd{$\pm$0.0} & 91.5\emptystd{$\pm$0.0} &   85.2\emptystd{$\pm$0.0} & 78.4 
  & 69.9\emptystd{$\pm$0.0} & 89.2\emptystd{$\pm$0.0} &   88.4\emptystd{$\pm$0.0} & 82.5\\
  & Keyphrase Clust. & \textbf{60.3}\std{$\pm$0.0} & \textbf{92.5}\std{$\pm$0.0} & \textbf{87.3}\std{$\pm$0.0} & \textbf{80.0} 
  & \textbf{72.3}\std{$\pm$0.0} & \textbf{90.2}\std{$\pm$0.0} & \textbf{90.0}\std{$\pm$0.0} & \textbf{84.2}\\

  \bottomrule
  \end{tabular}
  \caption{Comparing methods for integrating LLMs into entity canonicalization. ``CMVC'' refers to the multi-view clustering method of \citet{CMVC}, while ``KMeans'' refers to our simplified reimplementation of the same method. Where applicable, standard deviations are obtained by running clustering 5 times with different seeds.
  }
  \label{table:canonicalization_metrics_table} 
\end{table*}

\begin{table*}[t] 
  \centering
  \renewcommand{\arraystretch}{1}
  \fontsize{9}{9.5}\selectfont
  \setlength{\tabcolsep}{8pt}
  \begin{tabular}{@{} lr|rr|rr|rr @{} }
  \toprule\toprule
   &  \multirow{2}{*}{Dataset / Method} & \multicolumn{2}{c|}{Bank77} & \multicolumn{2}{c|}{CLINC} & \multicolumn{2}{c}{Tweet} \\  
   \cmidrule(lr){3-4} \cmidrule(lr){5-6} \cmidrule(lr){7-8}
   &  & \multicolumn{1}{c}{\textbf{Acc}} & \multicolumn{1}{c|}{\textbf{NMI}} & \multicolumn{1}{c}{\textbf{Acc}} & \multicolumn{1}{c|}{\textbf{NMI}}  & \multicolumn{1}{c}{\textbf{Acc}} & \multicolumn{1}{c}{\textbf{NMI}}  \\ 
   \midrule\midrule
  & SCCL  & --\emptystd{$\pm$0.0} & --\emptystd{$\pm$0.0} & --\emptystd{$\pm$0.0} & --\emptystd{$\pm$0.0} & 78.2\emptystd{$\pm$0.0} & 89.2\emptystd{$\pm$0.0}  \\
  & ClusterLLM  & 71.2\emptystd{$\pm$0.0} & --\emptystd{$\pm$0.0} & 83.8\emptystd{$\pm$0.0} & --\emptystd{$\pm$0.0} & --\emptystd{$\pm$0.0} & --\emptystd{$\pm$0.0}  \\
  \midrule
& KMeans  & 64.0\std{$\pm$0.0} & 81.7\std{$\pm$0.0} & 77.7\std{$\pm$0.0} & 91.5\std{$\pm$0.0}  & 57.5\std{$\pm$0.0} & 80.6\std{$\pm$0.0}  \\
\midrule
  \parbox[t]{1.5mm}{\multirow{3}{*}{\rotatebox[origin=c]{90}{\begin{small}ours\end{small}}}} & PCKMeans & 59.6\std{$\pm$0.0} & 79.6\std{$\pm$0.0} & \textbf{79.6}\std{$\pm$0.0} & 92.1\std{$\pm$0.0} & \textbf{65.3}\std{$\pm$0.0} & \textbf{85.1}\std{$\pm$0.0} \\
  & LLM Correction  & 64.1\emptystd{$\pm$0.0} & 81.9\emptystd{$\pm$0.0} & 77.8\emptystd{$\pm$0.0} & 91.3\emptystd{$\pm$0.0} & 59.0\emptystd{$\pm$0.0} & 81.5\emptystd{$\pm$0.0}   \\
  & Keyphrase Clustering & \textbf{65.3}\std{$\pm$0.0} & \textbf{82.4}\std{$\pm$0.0} & 79.4\std{$\pm$0.0} & \textbf{92.6}\std{$\pm$0.0} & 62.0\std{$\pm$0.0} & 83.8\std{$\pm$0.0}  \\
  \bottomrule
  \end{tabular}
  \caption{Comparing methods for integrating LLMs into text clustering. ``SCCL'' refers to \citet{SCCL} while ``ClusterLLM'' refers to \citet{Zhang2023ClusterLLMLL}. We use the same base encoders as those methods in our experiments. Where applicable, standard deviations are obtained by running clustering 5 times with different seeds.
  }
  \label{table:text_clustering_metrics_table} 
\end{table*}

\section{Results}
\subsection{Summary of Results} 
We summarize empirical results for entity canonicalization in \autoref{table:canonicalization_metrics_table} and text clustering in \autoref{table:text_clustering_metrics_table}.\footnote{
As discussed in Section \ref{sec:baseline}, when performing entity canonicalization, we assign mentions  to the same cluster if they contain the same entity surface form (e.g. ``\textit{Marseille}''), following prior work \citep{vashishth2018cesi, CMVC}. This approach leads to irreducible errors for polysemous noun phrases (e.g. ``Marseille'' may refer to the athletic club Olympique de Marseille or the city Marseille).

To our knowledge, we are the first to highlight the limitations of this ``surface form clustering'' approach. We present the optimal performance under this assumption in \autoref{table:canonicalization_metrics_table}, finding that the baseline of \citet{CMVC} is already near-optimal on some metrics, particularly for ReVerb45k.
} 
We find that using the LLM to expand textual representations is the most effective, achieving state-of-the-art results on both canonicalization datasets and significantly outperforming a K-Means baseline for all text clustering datasets. Pairwise constraint K-means, when provided with 20,000 pairwise constraints pseudo-labeled by an LLM, achieves strong performance on 3 of 5 datasets (beating the current state-of-the-art on OPIEC59k).   
Below, we conduct more in-depth analyses on what makes each method (in-)effective.

\subsection{LLMs excel at text expansion}
In \autoref{table:canonicalization_metrics_table} and \autoref{table:text_clustering_metrics_table}, we see that the ``Keyphrase Clustering'' approach is our strongest approach, achieving the best results on 3 of 5 datasets (and giving comparable performance to the next strongest method, pseudo-oracle PCKMeans, on the other 2 datasets). This suggests that LLMs are useful for expanding the contents of text to facilitate clustering.

What makes LLMs useful in this capacity? Is it the ability to specify task-specific modeling instructions, the ability to implicitly specify a similarity function via demonstrations, or do LLMs contain knowledge that smaller neural encoders lack?

\begin{table}[t] 
  \centering
  \renewcommand{\arraystretch}{1.1}
  \fontsize{9}{9.5}\selectfont
  \setlength{\tabcolsep}{2.5pt}
  \begin{tabular}{@{} r|c|ll @{} }
  \toprule
    \multirow{2}{*}{Dataset / Method} & OPIEC59k & \multicolumn{2}{c}{CLINC} \\  
    \cmidrule(lr){2-2} \cmidrule(lr){3-4} 
    & \multicolumn{1}{c|}{\textbf{Avg F1}} & \multicolumn{1}{c}{\textbf{Acc}} & \multicolumn{1}{c}{\textbf{NMI}} \\ 
   \midrule\midrule
Keyphrase Clust.  & \textbf{80.0} & \textbf{79.4}\std{$\pm$0.0} & \textbf{92.6}\std{$\pm$0.0}   \\
  w/o Instructions & 79.1 & 78.4\std{$\pm$0.0} & 92.7\std{$\pm$0.0} \\
  w/o Demonstrations  & 79.8 & 78.7\std{$\pm$0.0} & 91.8\std{$\pm$0.0}  \\
  \midrule
  Instructor-base & - - - & 74.8\std{$\pm$0.0} & 90.7\std{$\pm$0.0} \\
  Instructor-large & - - - & 77.7\std{$\pm$0.0} & 91.5\std{$\pm$0.0} \\
  Instructor-XL & - - - & 77.2\std{$\pm$0.0} & 91.9\std{$\pm$0.0} \\
  \citep{INSTRUCTOR} & & & \\
  \midrule
  Instructor-XL & - - - & 70.8\std{$\pm$0.0} & 88.6\std{$\pm$0.0}\\
  (GPT-3.5 prompt) & & & \\
  \bottomrule
  \end{tabular}
  \caption{We compare the effect of LLM intervention without demonstrations or without instructions. We see that GPT-3.5-based Keyphrase Clustering outperforms instruction-finetuned encoders of different sizes, even when we provide the same prompt.
  }
  \label{table:instruction_demonstration_ablation} 
\end{table}

We answer this question with an ablation study. For OPIEC59k and CLINC, we consider the ``Keyphrase Clustering'' technique but omit either the instruction or the demonstration examples from the prompt. For CLINC, we also compare with K-Means clustering on features from the Instructor model, which allows us to specify a short instruction to a small encoder.
We find empirically that providing either instructions or demonstrations in the prompt to the LLM enables the LLM to improve cluster quality, but that providing both gives the most consistent positive effect. Qualitatively, providing instructions but omitting demonstrations leads to a larger set of keyphrases with less consistency, while providing demonstrations without any instructions leads to a more focused group of keyphrases that sometimes fail to reflect the desired aspect (e.g. topic vs. intent).

Why is keyphrase clustering using GPT-3.5 in the instruction-only (``without demonstrations'') setting better than Instructor, which is an instruction-finetuned encoder?  While GPT-3.5's size is not published, GPT-3 contains 175B parameters, \texttt{Instructor-base/large/xl} contain 110M, 335M parameters, and 1.5B parameters, respectively. The modest scaling curve suggests that scale is not solely responsible.

Our prompts for Instructor are brief (e.g. ``Represent utterances for intent classification''), while our prompts for GPT-3.5 (in \autoref{sec:keyphrase_expansion_prompt}) are very  detailed. \texttt{Instructor-XL} does not handle long prompts well; in the bottom row of \autoref{table:instruction_demonstration_ablation}, we see that \texttt{Instructor-XL} performs poorly when given the same prompt that we give to GPT-3.5. We speculate that today's instruction-finetuned encoders are insufficient to support the detailed, task-specific prompts that facilitate few-shot clustering.

\subsection{The limitations of LLM post-correction}
LLM post-correction consistently provides small gains on datasets over all metrics -- between 0.1 and 5.2 absolute points of improvement. In \autoref{table:llm_correction}, we see that when we provide the top 500 most-uncertain cluster assignments to the LLM to reconsider, the LLM only reassigns points in a small minority of cases. Though the LLM pairwise oracle is usually accurate, the LLM is disproportionately inaccurate for points where the original clustering already had low confidence.

\begin{table}[t] 
  \centering
  \renewcommand{\arraystretch}{1.1}
  \fontsize{9}{9.5}\selectfont
  \setlength{\tabcolsep}{2.5pt}
  \begin{tabular}{@{} r|lll @{} }
  \toprule
    Dataset / Method &  OPIEC59k & CLINC & Tweet  \\  \midrule\midrule
  & \multicolumn{3}{c}{\textbf{Counts}} \\ 
   \midrule
Data Size  & 2138 & 4500 & 2472   \\
  \# of LLM Reassignmnts & 109 & 149 & 78 \\
  Accuracy of Reassignments  & 55.0 & 57.0 & 89.7  \\
    \midrule
Overall Accuracy of Pairwise & 86.7 & 95.0 & 96.8  \\
   Pseudo-Oracle & & & \\
  \bottomrule
  \end{tabular}
  \caption{When re-ranking the top 500 points in each dataset, the LLM rarely disagrees from the original clustering, and when it does, it is frequently wrong.}
  \label{table:llm_correction} 
\end{table}

\subsection{How much does LLM guidance cost?}
We've shown that using an LLM to guide the clustering process can improve cluster quality. However, large language models can be expensive; using a commercial LLM API during clustering imposes additional costs to the clustering process.

\begin{table}[t] 
  \centering
  \renewcommand{\arraystretch}{1.1}
  \fontsize{9}{9.5}\selectfont
  \setlength{\tabcolsep}{2.5pt}
  \begin{tabular}{@{} r|c|c|c|c @{} }
  \toprule
     &  & PCKMeans & Correction & Keyphrase  \\  \midrule\midrule
  \textbf{Method} & \textbf{Data Size} & \multicolumn{3}{c}{\textbf{Cost in USD}} \\ 
   \midrule
OPIEC59k  & 2138 & \$42.03 & \$12.73 & \$2.24  \\
  ReVerb45k & 12295 & \$33.81 & \$10.24 & \$10.66 \\
  Bank77  & 3080 & \$10.25 & \$3.38 & \$1.23  \\
  CLINC  & 4500 & \$9.77 & \$2.80 & \$0.95 \\
  Tweet  & 2472 & \$11.28 & \$3.72 & \$0.99  \\
  \bottomrule
  \end{tabular}
  \caption{We compare the pseudo-labeling costs of different LLM-guided clustering approaches. We used OpenAI's \texttt{gpt-3.5-turbo-0301} API in June 2023.
  }
  \label{table:clustering_cost} 
\end{table}

In \autoref{table:clustering_cost}, we summarize the pseudo-labeling cost of collecting LLM feedback using our three approaches. Among our three proposed approaches, pseudo-labeling pairwise constraints using an LLM (where the LLM must classify 20K pairs of points) incurs the greatest LLM API cost. While PCKMeans and LLM Correction both query the LLM the same number of times for each dataset, Keyphrase Correction's cost scales linearly with the size of the dataset, making this infeasible for clustering very large corpora.

\subsection{Using an LLM as a pseudo-oracle is cost-effective}
Using large language models increases the cost of clustering. Does the improved performance justify this cost? By employing a human expert to guide the clustering process instead of a large language model, could one achieve better results at a comparable cost?

\begin{figure}[t]
\centering
\begin{tikzpicture}
  \begin{axis}[
  title=Macro F1,
  width=\linewidth,
  line width=0.5,
  grid=major, 
  tick label style={font=\normalsize},
  legend style={nodes={scale=0.4, transform shape}},
  legend cell align={left},
  label style={font=\footnotesize},
  grid style={white},
  ylabel={OPIEC59k},
   y tick label style={
    font=\small
 },
 ymin=0.49,
 ymax=0.8,
 xmin=0,
 xmax=20000,
    x tick label style={font=\small, /pgf/number format/.cd,%
          scaled x ticks = false,
          set thousands separator={}},
    label style={font=\small},
    xticklabels={,0,,10K,,20K},
legend style={at={(0.0,0.65)}, anchor=south west,  draw=none, fill=none},
    height=3.5cm,
    width=4.1cm,
  ]
    \addplot[colorthree, line legend] coordinates
      {(0, 0.535) (1250, 0.530) (2500, 0.555) (5000, 0.566) (7500, 0.556)   (10000, 0.576) (15000, 0.5659)  (20000, 0.58654)};;

      \addplot+ [only marks, colorthree, mark=o, mark size=0.8pt, forget plot] coordinates
      {(0, 0.535) (1250, 0.536) (2500, 0.555) (5000, 0.566) (7500, 0.556)   (10000, 0.576) (15000, 0.5659)  (20000, 0.58654)};;

    \addplot[colorone, line legend] coordinates
      {(0, 0.535) (1250, 0.536) (2500, 0.545) (5000, 0.608) (7500, 0.662)   (10000, 0.688) (15000, 0.719)  (20000, 0.726)};;

      \addplot+ [only marks, colorone, mark=o, mark size=0.8pt, forget plot] coordinates
      {(0, 0.535) (1250, 0.536) (2500, 0.545) (5000, 0.608) (7500, 0.662)   (10000, 0.688) (15000, 0.719)  (20000, 0.726)};;
        
    \addplot [domain = 0:20000,
        thick,
        dashed,
        red, line legend
        ]{0.535};  

    \legend{PCKMeans (LLM oracle), PCKMeans (True oracle), KMeans}


  \end{axis}
\end{tikzpicture}
\hspace{-12pt}
\begin{tikzpicture}
  \begin{axis}[
  title=Pair F1,
  width=\linewidth,
  line width=0.5,
  grid=major, 
  tick label style={font=\normalsize},
  legend style={nodes={scale=0.4, transform shape}},
  label style={font=\normalsize},
  grid style={white},
   y tick label style={
    font=\small
 },
 ymin=0.8,
 ymax=0.95,
 xmin=0,
 xmax=20000,
    x tick label style={font=\small, /pgf/number format/.cd,%
          scaled x ticks = false,
          set thousands separator={}},
    label style={font=\small},
    xticklabels={,0,,10K,,20K},
legend style={at={(0.0,0.7)}, anchor=south west,  draw=none, fill=none},
    height=3.5cm,
    width=4.1cm,
  ]
    \addplot[colorthree, line legend] coordinates
      {(0, 0.856) (1250, 0.821) (2500, 0.815) (5000, 0.838) (7500, 0.830)   (10000, 0.828) (15000, 0.825)  (20000, 0.861)};;

      \addplot+ [only marks, colorthree, mark=o, mark size=0.8pt,forget plot] coordinates
      {(0, 0.856) (1250, 0.821) (2500, 0.815) (5000, 0.838) (7500, 0.830)   (10000, 0.828) (15000, 0.825)  (20000, 0.861)};;

    \addplot[colorone, line legend] coordinates
      {(0, 0.856) (1250, 0.833) (2500, 0.907) (5000, 0.912) (7500, 0.929)   (10000, 0.936) (15000, 0.938)  (20000, 0.942)};;

      \addplot+ [only marks, colorone, mark=o, mark size=0.8pt,forget plot] coordinates
      {(0, 0.856) (1250, 0.833) (2500, 0.907) (5000, 0.912) (7500, 0.929)   (10000, 0.936) (15000, 0.938)  (20000, 0.942)};;

        
    \addplot [domain = 0:20000,
        thick,
        dashed,
        red, line legend
        ]{0.856};  
      


  \end{axis}
\end{tikzpicture}

\begin{tikzpicture}
  \begin{axis}[
  width=\linewidth,
  line width=0.5,
  grid=major, 
  tick label style={font=\normalsize},
  legend style={nodes={scale=0.4, transform shape}},
  label style={font=\normalsize},
  grid style={white},
  ylabel={reverb45k},
  xlabel={\# of Constraints},
   y tick label style={
    font=\small
 },
 ymin=0.6,
 ymax=0.8,
 xmin=0,
 xmax=20000,
    x tick label style={font=\small, /pgf/number format/.cd,%
          scaled x ticks = false,
          set thousands separator={}},
    label style={font=\small},
    xticklabels={,0,,10K,,20K},
legend style={at={(0.0,0.65)}, anchor=south west,  draw=none, fill=none},
    height=3.5cm,
    width=4.1cm,
  ]
    \addplot[colorthree, line legend] coordinates
      {(0, 0.650) (1250, 0.660) (2500, 0.670) (5000, 0.680) (7500, 0.690)   (10000, 0.703) (15000, 0.7100)  (20000, 0.7201)};;

      \addplot+ [only marks, colorthree, mark=o, mark size=0.8pt, forget plot] coordinates
      {(0, 0.650) (1250, 0.660) (2500, 0.670) (5000, 0.680) (7500, 0.690)   (10000, 0.703) (15000, 0.7100)  (20000, 0.7201)};;

    \addplot[colorone, line legend] coordinates
      {(0, 0.650) (1250, 0.684) (2500, 0.680) (5000, 0.674) (7500, 0.684)   (10000, 0.709) (15000, 0.743)  (20000, 0.758)};;

      \addplot+ [only marks, colorone, mark=o, mark size=0.8pt, forget plot] coordinates
      {(0, 0.650) (1250, 0.684) (2500, 0.680) (5000, 0.674) (7500, 0.684)   (10000, 0.709) (15000, 0.743)  (20000, 0.758)};;

    \addplot [domain = 0:20000,
        thick,
        dashed,
        red, line legend
        ]{0.696};  


  \end{axis}
\end{tikzpicture}
\hspace{-12pt}
\begin{tikzpicture}
  \begin{axis}[
  width=\linewidth,
  line width=0.5,
  grid=major, 
  tick label style={font=\normalsize},
  legend style={nodes={scale=0.4, transform shape}},
  label style={font=\normalsize},
  grid style={white},
  xlabel={\# of Constraints},
   y tick label style={
    font=\small
 },
 ymin=0.75,
 ymax=1.0,
 xmin=0,
 xmax=20000,
    x tick label style={font=\small, /pgf/number format/.cd,%
          scaled x ticks = false,
          set thousands separator={}},
    label style={font=\small},
    xticklabels={,0,,10K,,20K},
legend style={at={(0.0,0.7)}, anchor=south west,  draw=none, fill=none},
    height=3.5cm,
    width=4.1cm,
  ]
    \addplot[colorthree, line legend] coordinates
      {(0, 0.866) (1250, 0.867) (2500, 0.868) (5000, 0.87) (7500, 0.873)   (10000, 0.856) (15000, 0.88)  (20000, 0.8855)};;

      \addplot+ [only marks, colorthree, mark=o, mark size=0.8pt,forget plot] coordinates
      {(0, 0.866) (1250, 0.867) (2500, 0.868) (5000, 0.87) (7500, 0.873)   (10000, 0.856) (15000, 0.88)  (20000, 0.8855)};;


    \addplot[colorone, line legend] coordinates
      {(0, 0.866) (1250, 0.885) (2500, 0.885) (5000, 0.884) (7500, 0.889)   (10000, 0.891) (15000, 0.894)  (20000, 0.904)};;

      \addplot+ [only marks, colorone, mark=o, mark size=0.8pt, forget plot] coordinates
      {(0, 0.866) (1250, 0.885) (2500, 0.885) (5000, 0.884) (7500, 0.889)   (10000, 0.891) (15000, 0.894)  (20000, 0.904)};;

    \addplot [domain = 0:20000,
        thick,
        dashed,
        red, line legend
        ]{0.891};  



  \end{axis}
\end{tikzpicture}
\vspace{-3em}
\caption{Collecting more pseudo-oracle feedback for pairwise constraint K-Means on OPIEC59k improves the Macro F1 metric without reducing other metrics. Compared to the same algorithm with true oracle constraints, we see the sensitivity of this algorithm to a noisy oracle.} 
\label{fig:opiec_pckmeans_feedback_curve}
\vspace{-0.8em}
\end{figure}
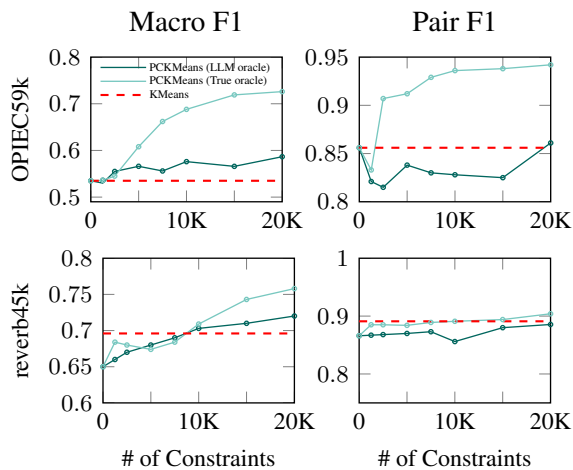

Since pseudo-labeling pairwise constraints requires the greatest API cost in our experiments, we take this approach as a case study. Given a sufficient amount of pseudo-oracle feedback, we see in \autoref{fig:opiec_pckmeans_feedback_curve} that pairwise constraint K-means is able to yield an improvement in Macro F1 (suggesting better purity of clusters) without dramatically reducing Pairwise or Micro F1.

Is this cost reasonable? For the \$41 spent on the OpenAI API for OPIEC59k (as shown in \autoref{table:clustering_cost}), one could hire a worker for 3.7 hours of labeling time, assuming an \$11-per-hour wage \citep{Hara2017ADA}. We observe that an annotator can label roughly 3 pairs per minute. Then, \$41 in worker wages would generate <700 human labels at the same cost as 20K GPT-3.5 labels.

Based on the feedback curve in \autoref{fig:opiec_pckmeans_feedback_curve}, we see that
GPT-3.5 is remarkably more effective than a true oracle pairwise constraint oracle at this price point; unless at least 2500 pairs labeled by a true oracle are provided, pairwise constraint KMeans fails to deliver any value for entity canonicalization. This suggests that if the goal is maximizing empirical performance, querying an LLM is more cost-effective than employing a human labeler.



\section{Conclusion}
We find that using LLMs in simple ways can provide consistent improvements to the quality of clusters for a variety of text clustering tasks. We find that LLMs are most consistently useful as a means of enriching document representations, and we believe that our simple proof-of-concept should motivate more elaborate approaches for document expansion via LLMs. 

\section{Acknowledgements}
This work was supported by a fellowship from NEC Research Laboratories. We are grateful to Wiem Ben Rim, Saujas Vaduguru, and Jill Fain Lehman for their guidance. We also thank Chenyang Zhao for providing valuable feedback on this work.



\bibliography{anthology,custom}
\bibliographystyle{acl_natbib}

\appendix

\section{Pairwise Constraint Pseudo-Oracle Prompt}
\label{sec:pairwise_constraint_prompt}
We use a task-specific prompt for the pairwise constraint pseudo-oracle, using 4 entities from each dataset as demonstration examples. We provide an example of one of the exact prompts used, for reference:

\paragraph{OPIEC59k Prompt}
\begin{quote}
    You are tasked with clustering entity strings based on whether they refer to the same Wikipedia article. To do this, you will be given pairs of entity names and asked if their anchor text, if used separately to link to a Wikipedia article, is likely referring to the same article. Entity names may be truncated, abbreviated, or ambiguous.

To help you make this determination, you will be given up to three context sentences from Wikipedia where the entity is used as anchor text for a hyperlink. Amongst each set of examples for a given entity, the entity for all three sentences is a link to the same article on Wikipedia. Based on these examples, you will decide whether the first entity and the second entity listed would likely link to the same Wikipedia article if used as separate anchor text.

Please note that the context sentences may not be representative of the entity's typical usage, but should aid in resolving the ambiguity of entities that have similar or overlapping meanings.

To avoid subjective decisions, the decision should be based on a strict set of criteria, such as whether the entities will generally be used in the same contexts, whether the context sentences mention the same topic, and whether the entities have the same domain and scope of meaning.

Your task will be considered successful if the entities are clustered into groups that consistently refer to the same Wikipedia articles.

1) B.A

Context Sentences:
a) "He matriculated at Jesus College , Oxford on 26 April 1616 , aged 16 , then transferred to Christ 's College , Cambridge ( B.A 1620 , M.A. 1623 , D.D. 1640 ) ."

2) M.D.

Context Sentence:
a) "David Satcher , M.D. , Ph.D. ."
b) "Livingston Farrand , M.D. , LL.D ."
c) "Dr. James Curtis Hepburn , M.D. , LL.D ."

Given this context, would B.A and M.D. link to the same entity's article on Wikipedia? No

1) B.A

Context Sentences:
a) "He matriculated at Jesus College , Oxford on 26 April 1616 , aged 16 , then transferred to Christ 's College , Cambridge ( B.A 1620 , M.A. 1623 , D.D. 1640 ) ."

2) bachelor

Context Sentence:
a) "He graduated in 1994 with a bachelor 's degree in sociology ."
b) "In 1976 , he graduated from Oneonta with a bachelor 's degree in sociology ."
c) "In 1991 , he graduated with bachelor 's degrees in linguistics and diplomacy ."

Given this context, would B.A and bachelor link to the same entity's article on Wikipedia? Yes

1) Duke of York

Context Sentences:
a) "In 1664 , the Duke of York , James , was granted this land by King Charles II ."
b) "Conflict continued concerning colonial limits until the Duke of York captured New Netherland in 1664 ."
c) "He served with the army under the Duke of York in Flanders as `` superintendent of Inland Navigation '' and won his confidence ."

2) Frederick

Context Sentence:
a) "The new settlement was named Frederick 's Town in honour of Prince Frederick , son of King George III and uncle of Queen Victoria ."
b) "The street was laid out during the 1820s , and takes its name from Prince Frederick , Duke of York and Albany , the younger brother of King George IV ."
c) "He entered the 1st Battalion of Grenadier Guards in 1795 and served in Holland under Prince Frederick , Duke of York and Albany , second son of George III ."

Given this context, would Duke of York and Frederick link to the same entity's article on Wikipedia? No

1) Academy Award

Context Sentences:
a) "Among her numerous accolades , Witherspoon has won an Academy Award , a BAFTA Award , and a Golden Globe , all for `` Walk the Line '' ."
b) "During her career , she has won an Academy Award , a BAFTA Award and was nominated for a Golden Globe for her role in `` The Last Picture Show '' ."
c) "Stone won an Academy Award , a BAFTA Award , and a Golden Globe Award for Best Actress for playing an aspiring actress in the highly successful musical film `` La La Land '' ( 2016 ) ."

2) Best Actor in Supporting Role

Context Sentence:
a) "The film was nominated for Best Actor in a Leading Role ( Oskar Werner ) , Best Actor in a Supporting Role ( Michael Dunn ) , Best Actress in a Leading Role ( Simone Signoret ) ."
b) "It was nominated for Best Actor in a Leading Role ( Warren Beatty ) , Best Actor in a Supporting Role ( Harvey Keitel and Ben Kingsley ) , Best Cinematography , Best Director , Best Music , Original Score , Best Picture and Best Writing , Screenplay Written Directly for the Screen ."
c) "The film was positively received by critics , and received seven Academy Award nominations , including Best Actor in a Leading Role ( Daniel Day-Lewis ) , Best Actor in a Supporting Role ( Pete Postlethwaite ) , Best Actress in a Supporting Role ( Emma Thompson ) , Best Director , and Best Picture ."

Given this context, would Academy Award and Best Actor in Supporting Role link to the same entity's article on Wikipedia? No

1) Justice Department

Context Sentences:
a) "In 1962 , he became the Justice Department 's first African-American lawyer in the Civil Rights Division ."
b) "Shortridge served as a special attorney for the Justice Department in Washington , D.C. from 1939 to 1943 ."
c) "The Justice Department appealed the decision to the Supreme Court in April 1992 , but the court declined to review the case ."

2) United States Department of State

Context Sentence:
a) "He spent 1946 -- 47 at the United States Department of State in Washington , D.C. as Palestine Desk Officer ."
b) "Atherton also served on the Philippine Commission and at the United States Department of State in Washington , D.C. ."
c) "The United States Department of State added Deif to its list of Specially Designated Global Terrorists on 8 September 2015 ."

Given this context, would Justice Department and United States Department of State link to the same entity's article on Wikipedia?
    
\end{quote}

\section{Keyphrase Expansion Prompt}
\label{sec:keyphrase_expansion_prompt}
We provide a domain-specific prompt to the keyword expansion clusterer, using 4 entities from each dataset as demonstration examples. We provide the exact prompts used for reference:

\paragraph{OPIEC59k Prompt}
\begin{quote}
I am trying to cluster entity strings on Wikipedia according to the Wikipedia article title they refer to. To help me with this, for a given entity name, please provide me with a comprehensive set of alternative names that could refer to the same entity. Entities may be weirdly truncated or ambiguous - e.g. "Wind" may refer to the band "Earth, Wind, and Fire" or to "rescue service". For each entity, I will provide you with a sentence where this entity is used to help you understand what this entity refers to. Generate a comprehensive set of alternate entity names as a JSON-formatted list.

Entity: "fictional character"

Context Sentences:
1) "Camille Raquin is a fictional character created by Émile Zola ."
2) "Druu is a fictional character appearing in comic books published by DC Comics ."
3) "Mallen is a fictional character that appears in comic books published by Marvel Comics .""

Alternate Entity Names: ["fictional characters", "characters", "character"]

Entity: "Catholicism"

Context Sentences:
1) "Years after Anne had herself converted , James avowed his Catholicism , which was a contributing factor to the Glorious Revolution ."
2) "The `` Catechism of the Catholic Church '' , representing Catholicism 's great regard for Thomism , the teachings of St. Thomas Aquinas , affirms that it is a Catholic doctrine that God 's existence can indeed be demonstrated by reason ."
3) "Palestinian Christians belong to one of a number of Christian denominations , including Eastern Orthodoxy , Oriental Orthodoxy , Catholicism ( Eastern and Western rites ) , Anglicanism , Lutheranism , other branches of Protestantism and others .""

Alternate Entity Names: ["Catholic Church", "Roman Catholic", "Catholic"]

Entity: "Wind"

Context Sentences:
1) "It was co-produced by Earth , Wind \& Fire 's keyboardist Larry Dunn ."
2) "Guitarist Tom Morello described the sound as `` Earth , Wind and Fire meets Led Zeppelin '' ."
3) "Taylor Mesplé also sang with Lampa along with Sydney Hostetler and the late Winston Ford ( Earth , Wind \& Fire , The Drifters ) .""

Alternate Entity Names: ["Earth \& Fire", "Earth", "Wind \& Fire"]

Entity: "Elizabeth"

Context Sentences:
1) "He also performed at the London Palladium for Queen Elizabeth ."
2) "On July 5 , 2010 , Ray was honored to be a host of the informal lunch for Queen Elizabeth 's visit to Toronto ."
3) "Its 1977 premiere was staged at the Royal Festival Hall in London as part of Queen Elizabeth 's Silver Jubilee .""

Alternate Entity Names: ["Elizabeth II", "HM"]

Entity: "Napoleon"

Context Sentences:
1) "Napoleon 's Imperial Guard is an example of this ."
2) "He studied at Paris until Napoleon 's return from Elba ."
3) "He said he had traveled to Egypt with Napoleon 's expedition .""

Alternate Entity Names:
["Napoleon Bonaparte", "Emperor Napoleon", "Napoleon I"]
\end{quote}

\end{document}